%% file: root.tex
\documentclass[letterpaper, 10 pt, conference]{ieeeconf}  % Comment this line out if you need a4paper

\usepackage[whole]{bxcjkjatype} % for Japanese

\IEEEoverridecommandlockouts                              % This command is only needed if 
\usepackage{graphics} % for pdf, bitmapped graphics files
\usepackage{epsfig} % for postscript graphics files
\usepackage{mathptmx} % assumes new font selection scheme installed
\usepackage{times} % assumes new font selection scheme installed
\usepackage{amsmath} % assumes amsmath package installed
\usepackage{amssymb}  % assumes amsmath package installed
\usepackage{multirow}
\usepackage[table,xcdraw]{xcolor}
\usepackage{threeparttable}
\usepackage{eucal}
\usepackage{balance}
\usepackage{mathrsfs}  
\usepackage{comment}
\usepackage[noend]{algpseudocode}
\usepackage{float}
\usepackage{textcomp}
\usepackage{mathcomp}
\usepackage{bm}
\usepackage{cite} 
\usepackage{booktabs}
\usepackage{url}
\usepackage{ifthen}

\usepackage{hyperref}

\def\eg{\textit{e.g}\onedot} 
\def\ie{\textit{i.e}\onedot}

\def\etal{\textit{et al}\onedot}

\makeatletter
\let\MYcaption\@makecaption
\makeatother

\usepackage[font=footnotesize]{subcaption}

\makeatletter
\let\@makecaption\MYcaption
\makeatother

\usepackage{subcaption}

\algnewcommand{\Break}{\textbf{break}}

\newcommand{\myvec}[1]{\mathbf{#1}}
\newcommand{\myset}[1]{\mathcal{#1}}

\newcommand{\mytag}[1]{\texttt{#1}}

\def\eg{{\it e.g.}}

\def\ie{{\it i.e.}}
\def\etal{{\it et al. }}

\newboolean{showchanges}
\setboolean{showchanges}{true} % true: 修正を表示, false: 通常版

\ifthenelse{\boolean{showchanges}}{
  
  \newcommand{\deleted}[1]{\textcolor{red}{\st{#1}}}
  
}{
  
  \newcommand{\deleted}[1]{}
  
}

\usepackage[colorinlistoftodos]{todonotes}

\title{\LARGE \bf
ReBound: Reset-Free Reinforcement Learning for Agile Driving\\via Reset-Aware Semi-Markov Bootstrapping
}

\author{Kohei Honda$^{1, 2, \dagger}$, Aiki Kada$^{1\dagger}$, Hirotaka Hosogaya$^{1}$% <-this % stops a space
\thanks{$^{\dagger}$Equal contribution. }% <-this % stops a space
\thanks{$^{1}$The Department of Mechanical Systems Engineering, Nagoya University, Aichi, Japan, {\tt\small honda.kohei.k3@f.mail.nagoya-u.ac.jp}}%
\thanks{$^{2}$CyberAgent AI Lab, Tokyo, Japan}%
\thanks{* This work was partially supported by JST ACT-X, Japan, under Grant Number JPMJAX24CN, and by the JST Moonshot R\&D Program, Japan, under Grant Number JPMJMS263D. }% <-this % stops a space
}

\begin{document}

% Teaser figure (full width) between the title block and the abstract
\makeatletter
\let\@oldmaketitle\@maketitle
\renewcommand{\@maketitle}{\@oldmaketitle
  \begin{center}
    \includegraphics[width=\textwidth]{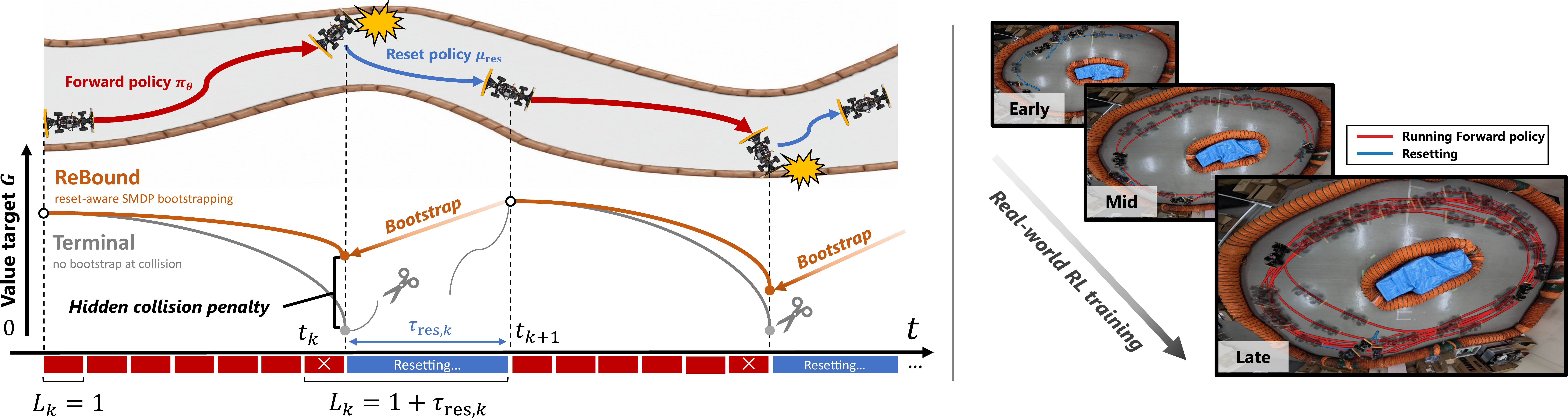}
    % ieeeconf typesets \@maketitle twice (once to measure it), so reset the counter to keep the teaser as Fig. 1.
    \setcounter{figure}{0}%
    \input{src/teaser.tex}
  \end{center}\vspace{-2mm}}
\makeatother

\maketitle

\thispagestyle{empty}
\pagestyle{empty}

%%%%%%%%%%%%%%%%%%%%%%%%%%%%%%%%%%%%%%%%%%%%%%%%%%%%%%%%%%%%%%%%%%%%%%%%%%%%%%%%

\begin{abstract}
\input{src/abstract.tex}
\end{abstract}
%%%%%%%%%%%%%%%%%%%%%%%%%%%%%%%%%%%%%%%%%%%%%%%%%%%%%%%%%%%%%%%%%%%%%%%%%%%%%%%%

% MAIN PARTS
\input{src/introduction.tex}
\input{src/related_work.tex}
\input{src/preliminaries.tex}
\input{src/method.tex}
\input{src/experiment.tex}
\input{src/conclusion.tex}

%%%%%%%%%%%%%%%%%%%%%%%%%%%%%%%%%%%%%%%%%%%%%%%%%%%%%%%%%%%%%%%%%%%%%%%%%%%%%%%%

% Appendices should appear before the acknowledgment.
% \input{acknowledgement}

\bibliographystyle{IEEEtran}
\bibliography{IEEEabrv, reference}

\end{document}

%% file: src/teaser.tex
\captionof{figure}{
    Overview of ReBound for value learning in reset-free reinforcement learning, which alternates between forward and reset policies.
    ReBound treats a collision-causing action and the subsequent recovery as a single semi-Markov transition and bootstraps from the post-recovery value, discounted by the actual recovery time, thereby aligning the value target with the reset-free objective.
    On a real vehicle, ReBound learns to drive substantially faster than conventional value-learning methods and model-based control.
}
\label{fig:overview}
\vspace{-3mm}

%% file: src/abstract.tex
We present ReBound (reset-aware semi-Markov bootstrapping), a value-learning method for reset-free reinforcement learning (RL) that learns agile driving through real-world RL without human intervention after training starts.
Reset-free RL explores by alternating between a forward policy that performs the task and a reset policy that restores a drivable state.
Standard reset-free RL treats collisions as episode terminations in value-function bootstrapping, truncating the return at each collision.
This conflicts with the reset-free objective, under which driving resumes after recovery, and acts as a \emph{hidden collision penalty} absent from the reward function.
ReBound instead applies semi-Markov bootstrapping: it treats the collision-causing action and the subsequent recovery as a single semi-Markov transition and bootstraps from the post-recovery value, discounted by the actual recovery time.
We further apply reward centering to stabilize value learning in the resulting continuing task.
In simulation and on a 1/10-scale vehicle, ReBound substantially outperforms conventional value-learning methods and model-based control.

%% file: src/introduction.tex
\section{Introduction}
\label{sec:introduction}

Reinforcement learning (RL) has outperformed top human drivers in high-fidelity racing simulators~\cite{fuchs2021super,wurman2022outracing}
and has beaten world champions in real-world drone racing~\cite{kaufmann2023champion}.
However, autonomous agile driving in real cars, even 1/10-scale vehicles, still falls short of skilled human performance.
A central obstacle is the sim-to-real gap.
At high speed, tires operate near the friction limit, and tire slip, suspension dynamics, and actuation delays dominate the vehicle's behavior.
These phenomena are hard to model accurately, so policies trained only in simulation do not transfer readily to real vehicles.

We therefore aim to learn agile driving for 1/10-scale vehicles through real-world RL, \ie, learning directly in the physical world.
Our vision is simple: place the vehicle on a track, start training, and let it practice autonomously through trial and error until it drives fast a few hours later (Fig.~\ref{fig:overview}, right).
This requires a framework that continues collecting data and learning without human assistance, regardless of how often the vehicle fails.
We therefore adopt reset-free RL~\cite{eysenbach2018leave,sharma2021autonomous,sharmaautonomous},
which explores by switching between a forward policy that maximizes the task return and a reset policy that brings the vehicle back to a drivable state after a collision.

Applying reset-free RL to agile driving poses its own difficulties.
First, successful agile driving lies close to the failure boundary.
The fastest racing line passes close to the walls, and the maximum cornering speed is just below the speed at which the vehicle crashes.
Learning thus hinges on accurate value estimates near the collision boundary.
Second, in reset-free RL, an episode-ending event, \ie, a collision, does not end the overall task.
Most collisions involving a small vehicle allow rapid recovery: once the reset policy brings the vehicle back to a drivable state within a few seconds, driving resumes.
Nevertheless, collisions are commonly treated as episode terminations: the post-collision value is not bootstrapped, as if all value obtainable after recovery were lost~\cite{hildisch2025drive,wang2026immediate}.
As we show, this target is misaligned with the reset-free objective and introduces an implicit additional collision penalty that is not specified by the reward function.
We call this the \emph{hidden collision penalty}.
Counterintuitively, this gap grows as recovery becomes faster: the faster the reset policy recovers the vehicle, the further the learned value deviates from the reset-free objective.

To address this, we present ReBound (reset-aware semi-Markov bootstrapping), which applies semi-Markov decision process (SMDP) bootstrapping~\cite{bradtke1994smdp,sutton1999options} to reset-free RL.
ReBound treats the collision-causing forward action and the subsequent recovery by a fixed reset policy as a single SMDP macro-transition
and bootstraps from the post-recovery value, discounted by the actual recovery time.
The value update thus treats a collision as a recoverable event rather than the end of the task, consistent with the reset-free setting.
To stabilize value learning in the resulting continuing task, we further combine ReBound with reward centering (RC)~\cite{naik2024reward} extended to variable-length transitions.
Because ReBound changes only the critic's learning target, it can be readily integrated into existing off-policy RL algorithms.

Our contributions are as follows:
\begin{itemize}
\item We identify a mismatch between the reset-free objective and the conventional collision-terminal update, and show analytically that the resulting gap, the hidden collision penalty, grows as recovery becomes faster.
      To resolve it, we present ReBound, which bootstraps across recovery using a semi-Markov transition to maintain accurate value estimates near the collision boundary.
\item We apply ReBound to agile driving with a 1/10-scale vehicle. In simulation, it learns to drive faster than both methods that use conventional value updates and a model-based controller, while also reducing value-estimation error near collisions. Through real-world RL on the physical vehicle, it further learns to drive faster than existing methods, regardless of whether manual resets are used.
\end{itemize}

%% file: src/related_work.tex
\section{Related Work}
\label{sec:related}

\subsection{Control and Learning for Agile Driving}

For agile driving with real 1/10-scale vehicles such as RoboRacer~\cite{o2020f1tenth}, online optimization methods such as model predictive control (MPC) are still widely used~\cite{honda2024stein}.
However, their performance degrades in high-speed, low-friction regimes where the dynamics are hard to model, and their finite-horizon optimization relies on predefined maps and localization for long-term strategies.

RL can overcome these limitations. TC-Driver~\cite{ghignone2024tc} and WheeledLab~\cite{han2025wheeled} transfer policies from simulation to reality, but the large sim-to-real gap for vehicles makes such transfer difficult.
Several studies have therefore investigated direct real-world learning.
Evans \etal performed online learning on a real vehicle under a model-based supervisor that guarantees safety~\cite{evans2023safe,evans2023bypassing}.
FastRLAP~\cite{stachowicz2023fastrlap} adapts in the real world using pretrained visual representations and checkpoints.
Hildisch \etal performed onboard training with a baseline controller that returns the vehicle to a reference path after events such as leaving the track~\cite{hildisch2025drive}.
All of these methods rely on state estimation, including localization.

\subsection{Reset-Free RL}

In real-world RL, failures such as collisions interrupt task execution and exploration, and unlike in simulation, the environment cannot simply be reinitialized.
Reset-free RL automates recovery by alternating between a forward policy and a reset policy, reducing manual resets~\cite{eysenbach2018leave,sharma2021autonomous,sharmaautonomous}.
It has been applied to many tasks on physical robots, including legged locomotion~\cite{yang2022safe}, cable insertion~\cite{luo2024serl},
dexterous manipulation~\cite{gupta2021reset,zhuingredients}, and mobile manipulation~\cite{sun2022fully}.
Prior work has mainly focused on learning reset or recovery policies and deciding when to switch between them and the forward policy; when learned jointly, such auxiliary policies can also shape the forward policy's curriculum~\cite{eysenbach2018leave,gupta2021reset,patil2024intelligent}.
Related work has also used recovery policies to reduce the frequency of failures by intervening before unsafe outcomes~\cite{thananjeyan2021recovery,daneshmand2026safeexplorer}.
Despite substantial work on how to execute and trigger such auxiliary policies, little attention has been paid to how their execution should be reflected in the forward policy's value learning after a collision.

Most methods treat the switch as a termination, which, as we show, discards the post-collision value at every collision.
Wang \etal also question resetting the environment at every collision,~\cite{wang2026immediate} but collisions remain local terminations in their value learning, so without global resets, as in our setting, their scheme reduces to the Terminal update.
RISC~\cite{patil2024intelligent} instead treats the switch as a time-out and continues bootstrapping with the forward agent's own value function.
In agile driving, however, this ignores the time spent in recovery, so the post-collision value is overestimated and learning becomes unstable.
ReBound lies between these two extremes, \emph{terminal} and \emph{timeout}: with a fixed reset policy, it bootstraps from the state after recovery, discounted by the actual recovery time.

%% file: src/preliminaries.tex
\section{Preliminaries}
\label{sec:preliminary}

\subsection{Formulation of Localization-Free Agile Driving}
\label{subsec:formulation-agile-driving}

We target high-speed 1/10-scale autonomous vehicles equipped with a 2D LiDAR, such as RoboRacer~\cite{o2020f1tenth}.
Conventional model-based controllers for such vehicles use LiDAR-based localization to estimate the vehicle state for planning~\cite{honda2024stein}.
At high speed, however, tire slip and unmodeled dynamics can cause large localization errors, and collisions during real-world exploration make reliable localization even harder to assume.
We therefore learn a policy that drives without localization, using only onboard local sensors.

\paragraph{Action space}
The action consists of the commanded speed $\hat{v}_t$ and steering angle $\hat{\delta}_t$, each normalized by its predefined maximum:
$
\myvec{a}_t
=
\begin{bmatrix}
    \hat{v}_t, &
    \hat{\delta}_t
\end{bmatrix}^{\top}
\in \myset{A}
= [-1,1]^2.
$

\paragraph{State space}
Since a single observation from the 2D LiDAR, IMU, and wheel encoders cannot capture the vehicle's dynamic state, the state concatenates the last $N_{\mathrm{h}}$ observations:
% \begin{align}
$
\myvec{s}_t
    =
    [
        \myvec{o}_{t-N_{\mathrm{h}}+1},\;
        \dots,\;
        \myvec{o}_{t}
    ]^{\top}
    \in \myset{S},
    % \label{eq:state}
    % \\
    \myvec{o}_t
    =
    [
        v_t,\;
        \omega_t,\;
        \Delta v_t,\;
        \Delta\omega_t,\;
        \myvec{a}_{t-1},\;
        \myvec{u}_{t-1},\;
        \myvec{d}_t
    ]^{\top},
$
% \end{align}
where
$v_t$ is the vehicle speed measured by the wheel encoders,
$\omega_t$ is the yaw rate measured by the IMU,
$\Delta v_t=v_t-v_{t-1}$ and $\Delta\omega_t=\omega_t-\omega_{t-1}$ are their one-step differences,
$\myvec{a}_{t-1}$ is the previous action of the forward policy,
$\myvec{u}_{t-1}$ is the command actually sent to the vehicle from the motor driver,
and $\myvec{d}_t$ is the vector of 2D LiDAR ranges.
Each quantity is normalized by a predefined scale, and steering angles are encoded by their sine and cosine.

\paragraph{Reward function}
The reward encourages fast forward driving while penalizing high-speed collisions.
Following~\cite{fuchs2021super}, we define
\begin{align}
    r_t(\myvec{s}_t,\myvec{a}_t)
    =
    v_{t+1} - \lambda_{\mathrm{col}}\, c_t \left| v_t \right|,
    \label{eq:agile-driving-reward}
\end{align}
where $v_{t}\in[-1,1]$ is the normalized speed,
$c_t = \mathbb{I}\left\{ \myvec{s}_{t+1} \in \myset{C} \right\}$ is the collision indicator,
$\myset{C}\subset\myset{S}$ is the set of collision states,
and $\lambda_{\mathrm{col}}>0$ weights the collision penalty, which is proportional to the speed just before the collision.

\subsection{Q-Learning and Termination at Collisions}
\label{subsec:collision-termination-qlearning}

To illustrate how termination affects value learning, consider standard one-step Q-learning as used in off-policy RL.
We assume that a collision terminates the episode, as is common in episodic RL.
The critic loss and target are
\begin{align}
    \mathcal{L}^{\mathrm{term}}_Q(\theta)
    &=
    \mathbb{E}_{
        (\myvec{s}_t,\myvec{a}_t,r_t,\myvec{s}_{t+1},c_t)
        \sim \myset{B}
    }
    \left[
        \left(
            Q_\theta(\myvec{s}_t,\myvec{a}_t)
            -
            y_t
        \right)^2
    \right],
    \label{eq:critic-loss}
    \\
    y_t
    &=
    r_t
    +
    \gamma(1-c_t)
    \max_{\myvec{a}'\in\myset{A}}
    Q_{\bar{\theta}}
    \left(
        \myvec{s}_{t+1},
        \myvec{a}'
    \right),
    \label{eq:mdp-target}
\end{align}
where $\myset{B}$ is the replay buffer, $\gamma\in[0,1)$ is the discount factor,
and the collision indicator $c_t$ in~\eqref{eq:agile-driving-reward} serves as the termination flag.
For a collision transition, $c_t=1$ and $y_t=r_t$, so the post-collision value is not bootstrapped.
For a policy $\pi$, \eqref{eq:mdp-target} thus estimates the expected discounted return truncated at the first collision,
\begin{align}
    Q^{\mathrm{term}}(\myvec{s}_t,\myvec{a}_t)
    =
    \mathbb{E}_{\pi}\left[
        G_t^{\mathrm{term}}
        \,\middle|\,
        \myvec{s}_t,\myvec{a}_t
    \right], \quad
    G_t^{\mathrm{term}}
    =
    \sum_{i=0}^{t_{\mathrm{c}}-t}\gamma^{i}\,r_{t+i},
    \label{eq:q-term}
\end{align}
where $t_{\mathrm{c}}=\min\{t'\geq t : c_{t'}=1\}$ is the time of the first collision-causing action at or after $t$.

%% file: src/method.tex
\section{Reset-Free Agile Driving via ReBound}
\label{sec:method}

To learn the forward policy $\pi$ through real-world RL, we build a reset-free system with a model-based reset policy (Fig.~\ref{fig:overview}) and present ReBound, which uses semi-Markov bootstrapping across the reset interval.

\subsection{Reset-Free RL for Agile Driving}
\label{subsec:reset-free}

Following the reset-free RL framework~\cite{eysenbach2018leave,sharma2021autonomous,sharmaautonomous}, we design the controller as a hybrid system with a \emph{Forward} mode that performs the task and a \emph{Reset} mode that recovers the vehicle after a collision.
Assuming that the state includes the history needed for mode switching, we define the composite policy as
\begin{align}
    \Pi_{\theta}
    \left(
        \myvec{a}_t \mid \myvec{s}_t
    \right)
    =
    \begin{cases}
        \pi_{\theta}
        \left(
            \myvec{a}_t \mid \myvec{s}_t
        \right),
        & \text{if }
        \{\myvec{s}_{t-j}\}_{j=0}^{\tau_{\mathrm{m}}}
        \cap \myset{C} = \emptyset,\\
        \mu_{\mathrm{res}}
        \left(
            \myvec{a}_t \mid \myvec{s}_t
        \right),
        & \text{otherwise},
    \end{cases}
    \label{eq:comp-policy}
\end{align}
where $\mu_{\mathrm{res}}$ is the reset policy and $\tau_{\mathrm{m}}$ is a margin that prevents chattering:
control returns to the forward policy only after the vehicle has remained collision-free for $\tau_{\mathrm{m}}$ steps under the reset policy.

Because low-speed vehicle behavior is relatively easy to model, model-based control can reliably recover the vehicle as long as neither the track nor the vehicle is damaged by collisions.
We therefore implement the reset policy as a fixed model predictive path integral (MPPI) controller~\cite{williams2018information},
which solves the following optimization problem over a prediction horizon $T$ by Monte Carlo sampling:
\begin{align}
    \mu_{\mathrm{res}}
    =
    \operatorname*{arg\,max}_{\mu}
    \left\{
        \mathbb{E}_{\mu}
        \left[
            \hat{r}_{t:t+T-1}
        \right]
        -
        \lambda_{\mathrm{res}}
        \mathbb{D}_{\mathrm{KL}}
        \left(
            \mu
            \parallel
            \mu^{\mathrm{prior}}
        \right)
    \right\},
    \label{eq:mppi}
\end{align}
where $\mu$ is a Gaussian policy, $\mu^{\mathrm{prior}}$ is the distribution obtained at the previous optimization step,
$\hat{r}_{t:t+T-1}$ is the cumulative reward defined by~\eqref{eq:agile-driving-reward}, predicted by the model, and $\lambda_{\mathrm{res}}$ is the temperature parameter.

\subsection{ReBound: Reset-Aware Semi-Markov Bootstrapping}
\label{subsec:rebound}

Under the composite policy in~\eqref{eq:comp-policy}, the cycle Forward $\rightarrow$ Reset $\rightarrow$ Forward $\rightarrow\cdots$ repeats as shown in Fig.~\ref{fig:overview} (left), enabling exploration and learning on the real vehicle indefinitely.
The collision-terminal update in Section~\ref{subsec:collision-termination-qlearning} sets the post-collision value to zero, yet in our setting driving always resumes after a recoverable collision.
Because a collision does not end the task, ReBound learns the continuing value of the composite policy, including recovery.
Because only the forward policy is learned, we use forward-action steps as decision points and construct an embedded semi-Markov decision process (SMDP)~\cite{bradtke1994smdp,sutton1999options} that incorporates the intervening fixed reset control.

\subsubsection{SMDP Including the Reset Policy}

An SMDP extends an MDP to allow variable intervals between decisions.
As shown in Fig.~\ref{fig:overview}, let $t_k$ be the time of the $k$-th forward action and $L_k=t_{k+1}-t_k$ the number of steps until the forward policy acts next;
we write $\myvec{s}_k=\myvec{s}_{t_k}$ and $\myvec{a}_k=\myvec{a}_{t_k}$.
Thus, $L_k=1$ during forward driving.
If the forward action at $t_k$ causes a collision and recovery requires $\tau_{\mathrm{res},k}$ reset steps, then $L_k=1+\tau_{\mathrm{res},k}$.

Forward actions receive the task reward~\eqref{eq:agile-driving-reward}, including the penalty for any collision-causing action, and reset actions receive zero reward.
The reset policy is a fixed external controller outside the learned task; rewarding its actions using~\eqref{eq:agile-driving-reward} would penalize the forward policy for the reset controller's behavior (\eg, reversing away from a wall yields a negative speed reward).
The cost of a collision is thus the collision penalty plus the actual time lost during recovery, expressed through discounting.
The discounted transition reward at $t_k$ is therefore
% \begin{align}
$
R_k^\gamma
    =
    \sum_{j=0}^{L_k-1}\gamma^j r_{t_k+j}
    =
    r_{t_k}.
    % \label{eq:macro-reward}
$
% \end{align}

Each SMDP transition is therefore represented by the tuple
$
    \left(
        \myvec{s}_k,\,
        \myvec{a}_k,\,
        R_k^\gamma,\,
        L_k,\,
        \myvec{s}_{k+1}
    \right),
$
where $\myvec{s}_{k+1}=\myvec{s}_{t_{k+1}}$ is the state at which the forward policy acts next, \ie, the state after recovery if the transition contains a reset interval.
Based on these transitions, we learn the Q-function using
\begin{align}
    \mathcal{L}^{\mathrm{SMDP}}_Q(\theta)
    &=
    \mathbb{E}_{
        (
            \myvec{s}_k,\myvec{a}_k,
            R_k^\gamma,L_k,\myvec{s}_{k+1}
        )
        \sim \myset{B}
    }
    \left[
        \left(
            Q_\theta(\myvec{s}_k,\myvec{a}_k)
            -
            y_k
        \right)^2
    \right],
    \label{eq:critic-loss-smdp}
    \\
    y_k
    &=
    R_k^\gamma
    +
    \gamma^{L_k}
    \max_{\myvec{a}'\in\myset{A}}
    Q_{\bar{\theta}}
    \left(
        \myvec{s}_{k+1},\myvec{a}'
    \right).
    \label{eq:smdp-target}
\end{align}
Unlike~\eqref{eq:mdp-target}, this update uses no termination flag because no true terminal state exists.\footnote{Stopping data collection, \eg, at the end of an experiment, is treated as a truncation rather than a termination~\cite{pardo2018time}.}

The target of reset-free learning is therefore the sum of rewards under the composite policy, discounted in real time,
\begin{align}
    G_t
    =
    \sum_{i=0}^{\infty}\gamma^{i}\,r_{t+i}.
    \label{eq:return-rf}
\end{align}
% This is the same position as the argument that time-outs should be bootstrapped rather than treated as terminations~\cite{pardo2018time}.
Since $t_{k+1}=t_k+L_k$, $G_{t_k}=\sum_{j=0}^{\infty}\gamma^{t_{k+j}-t_k}R_{k+j}^\gamma$.
Hence, for policy evaluation (replacing the maximization with the forward policy's action), the fixed point of~\eqref{eq:smdp-target} is
\begin{align}
    Q^{\Pi}(\myvec{s},\myvec{a})
    =
    \mathbb{E}_{\Pi}\left[
        G_t
        \,\middle|\,
        \myvec{s}_t=\myvec{s},\
        \myvec{a}_t=\myvec{a}
    \right],
    \label{eq:q-rebound-fixed}
\end{align}
where $t$ is a forward-policy decision time; that is, ReBound learns the continuing objective defined by the task reward, with zero reward during the externally controlled reset intervals.

\subsubsection{Comparison with the Collision-Terminal Value Update}
\label{subsec:terminal-comparison}
We compare the ReBound value $Q^{\Pi}$~\eqref{eq:q-rebound-fixed} with the collision-terminal value $Q^{\mathrm{term}}$~\eqref{eq:q-term}.
Let $\tau_{\mathrm{c}}=t_{\mathrm{c}}-t+1$ be the number of steps up to and including the collision-causing step,
and $t_{\mathrm{rec}}=t+\tau_{\mathrm{c}}+\tau_{\mathrm{res}}$ the time at which recovery completes.
Because the reward is zero during recovery, $G_t=G_t^{\mathrm{term}}+\gamma^{\tau_{\mathrm{c}}+\tau_{\mathrm{res}}}G_{t_{\mathrm{rec}}}$ holds, which gives
\begin{align}
    &Q^{\Pi}(\myvec{s},\myvec{a})-Q^{\mathrm{term}}(\myvec{s},\myvec{a})
    = \mathbb{E}_{\Pi}\left[
        \gamma^{\tau_{\mathrm{c}}+\tau_{\mathrm{res}}}\,
        V^{\Pi}(\myvec{s}_{t_{\mathrm{rec}}})
        \,\middle|\,
        \myvec{s},\
        \myvec{a}
    \right],
    \label{eq:terminal-bias}
\end{align}
where $V^{\Pi}$ is the state value of the composite policy.
The right-hand side of~\eqref{eq:terminal-bias} quantifies the mismatch: relative to the reset-free objective, the collision-terminal target additionally discards all value obtainable after recovery.
We call this gap the hidden collision penalty because it acts as a collision cost that is not specified by the reward function.
The gap is larger near collisions and with faster recovery, and varies during learning with the policy's own value.

How much, then, should a collision cost under the reset-free objective?
As a reference, consider a hypothetical collision from which the vehicle recovers instantly ($\tau_{\mathrm{res}}=0$).
Its return is $G_t^{\mathrm{term}}+\gamma^{\tau_{\mathrm{c}}}G_{t_{\mathrm{rec}}}$, so its only cost is the collision penalty contained in $G_t^{\mathrm{term}}$.
Relative to this reference, the additional value each target loses due to a collision is
\begin{align}
    C^{\mathrm{term}}
    =
    \gamma^{\tau_{\mathrm{c}}}G_{t_{\mathrm{rec}}},
    \qquad
    C^{\mathrm{ReBound}}
    =
    \left(1-\gamma^{\tau_{\mathrm{res}}}\right)\gamma^{\tau_{\mathrm{c}}}G_{t_{\mathrm{rec}}}.
    \label{eq:collision-cost}
\end{align}
Whereas the collision-terminal target loses the entire post-recovery value, ReBound loses only the portion attributable to delaying that value by $\tau_{\mathrm{res}}$ steps, \ie, the opportunity cost of the time spent in recovery.
The ReBound target therefore exhibits only a shallow dip at the collision boundary (Fig.~\ref{fig:overview}, left).
In contrast, discounting the reset interval as a single step ($L_k=1$), as in the time-out treatment of the policy switch~\cite{patil2024intelligent}, ignores even this opportunity cost, effectively treating recovery as instantaneous.
We examine this timeout scheme in Section~\ref{subsec:sim-performance}.

\subsubsection{Stabilizing Learning with Reward Centering}
\label{subsec:rc}

Because ReBound continues bootstrapping after collisions, the critic estimates a long-horizon discounted return in a continuing task.
As $\gamma$ approaches 1, the Q-values contain a large common component arising from the average reward, which can destabilize learning.
We therefore introduce reward centering (RC)~\cite{naik2024reward}, which subtracts a per-step baseline $\bar{r}$ from every reward:
% \begin{align}
$
    \widetilde{R}_k^\gamma
    =
    \sum_{j=0}^{L_k-1}
    \gamma^j
    \left(
        r_{t_k+j}-\bar{r}
    \right)
    =
    r_{t_k}-W_k^\gamma\bar{r},
    % \label{eq:centered-macro-reward}
$
where $W_k^\gamma=(1-\gamma^{L_k})/(1-\gamma)$ is the centering weight for variable-length transitions ($W_k^\gamma=1$ for $L_k=1$, recovering standard RC).
Following value-based RC~\cite{naik2024reward}, $\bar{r}$ is updated together with the critic using the minibatch TD error.
ReBound thus learns the Q-function with
\begin{align}
    \mathcal{L}^{\mathrm{ReBound}}_Q(\theta)
    &=
    \mathbb{E}_{
        (
            \myvec{s}_k,\myvec{a}_k,
            \widetilde{R}_k^\gamma,L_k,\myvec{s}_{k+1}
        )
        \sim \myset{B}
    }
    \left[
        \left(
            Q_\theta(\myvec{s}_k,\myvec{a}_k)
            -
            y_k
        \right)^2
    \right],
    \label{eq:critic-loss-rebound}
    \\
    y_k
    &=
    r_{t_k}
    -
    W_k^\gamma\bar{r}
    +
    \gamma^{L_k}
    \max_{\myvec{a}'\in\myset{A}}
    Q_{\bar{\theta}}
    \left(
        \myvec{s}_{k+1},\myvec{a}'
    \right).
    \label{eq:critic-target-rebound}
\end{align}
Let $\widetilde{Q}$ denote the Q-function for centered rewards estimated by this critic.
For a fixed $\bar{r}$, comparing~\eqref{eq:critic-target-rebound} with~\eqref{eq:smdp-target} shows that, regardless of $L_k$,
\begin{align}
    \widetilde{Q}(\myvec{s},\myvec{a})
    =
    Q^{\Pi}(\myvec{s},\myvec{a})
    -
    \frac{\bar{r}}{1-\gamma}.
    \label{eq:rc-shift}
\end{align}
With $W_k^\gamma$, RC shifts $Q^{\Pi}$ by a constant, stabilizing critic learning while preserving the value-function structure.

%% file: src/experiment.tex
\section{Experiments}
\label{sec:experiments}

We investigate the following questions:
(Q1)~Does ReBound improve driving performance in reset-free learning?
(Q2)~How does it compare with time-out and terminal bootstrapping, and how much do SMDP bootstrapping and reward centering each contribute?
(Q3)~Does it learn the value predicted in Section~\ref{subsec:terminal-comparison}, and how does its effect depend on the recovery time?
(Q4)~Can the real vehicle learn high-speed driving without manual resets?

\subsection{Experimental Setup}
\label{subsec:experimental_setup}

\subsubsection{Vehicle and Environments}
In the real-world experiments, we use a 1/10-scale RoboRacer~\cite{o2020f1tenth} vehicle equipped with a 2D LiDAR (Hokuyo UTM-30LX), an IMU, and wheel encoders (VESC).
Both learning and control run on an onboard Intel NUC mini PC equipped with an Intel Core i7 CPU, without GPU acceleration, and the control frequency is 20,Hz.
The indoor courses are enclosed loops bounded by duct hoses, so the vehicle cannot physically leave them (Fig.~\ref{fig:real-eval-traj}).
We prepared two courses: the narrow \mytag{LabTrack} (centerline lap length about 13\,m, average width 2\,m) and the wider \mytag{LabCircuit} (about 40\,m, 2.2\,m).
The courses have different, very slippery floor materials.
We set a speed limit of 3.0\,m/s on both courses.

For simulation, we used F1TENTH gym~\cite{o2020f1tenth} with the same sensors as the real vehicle and dynamics based on the dynamic bicycle model (DBM)~\cite{rajamani2006vehicle}.
We prepared three simulated courses (Fig.~\ref{fig:sim_trajectories}).
\mytag{LabTrack-Sim} is reconstructed from a 3D scan of the real \mytag{LabTrack} environment.
\mytag{MoscowRaceway} and \mytag{Oschersleben} are scaled-down versions of real circuits, with a lap length of about 300\,m and a width of about 2.2\,m.
The friction coefficient was set to 0.4 to simulate a slippery surface, and the speed limit was 3.0\,m/s on \mytag{LabTrack-Sim} and 8\,m/s on the circuits.

\subsubsection{Training Procedure}
As described in Section~\ref{subsec:formulation-agile-driving}, the vehicle learns high-speed driving using only LiDAR, IMU, and wheel-encoder data, without localization or maps.
In the real world, the only manual operations are placing the vehicle on the course and swapping batteries; all learning and control are autonomous.
With onboard computation, gradient updates were batched rather than performed in real time: the vehicle paused after each collision recovery or after 500 steps to update the model.
Defining each data-collection period before gradient updates as one episode, we trained for up to 500 episodes in simulation, 900 episodes on \mytag{LabTrack}, and 600 episodes on \mytag{LabCircuit}.
We used 5 seeds in simulation and 1 seed on each real course.

\subsubsection{Evaluation Setup}
After training on each course, the final policy for each seed was run without exploration noise from 10 equally spaced start positions on that course.
An evaluation episode ends after 500 steps or at the first collision.
The metrics are the sum of rewards~\eqref{eq:agile-driving-reward} over an episode (Episode Return) and the fraction of episodes ending in a collision (Collision Rate).
This protocol is deliberately stricter than the training objective: it evaluates only the forward policy, giving no credit for post-recovery driving.
It enables a fair comparison with MPPI and reflects deployment, where any collision is a failure.
Performance with the reset policy in the continuing setting is shown by the training-return curves and real-world evaluation (Table~\ref{tab:real-world-evaluation}).

\subsubsection{Compared Methods}
\begin{itemize}
    \item \textbf{MPPI}: A model-based controller without learning that optimizes the same reward as RL using the kinematic bicycle model (KBM)~\cite{rajamani2006vehicle}\footnote{The KBM cannot account for slip, but MPPI with the DBM was unstable at low speed~\cite{aoki2021comparative}.} and a LiDAR-based local occupancy grid, implemented as in~\cite{honda2025model}.
    \item \textbf{Terminal-TD-MPC2}: The critic of TD-MPC2~\cite{hansen2024tdmpc2}, a state-of-the-art off-policy RL method, is trained using the standard collision-terminal update~\eqref{eq:mdp-target}.
    \item \textbf{Timeout-TD-MPC2}: Following the time-out treatment of RISC~\cite{patil2024intelligent}, the critic bootstraps from the state after recovery with a one-step discount $\gamma$ ($L_k=1$), \ie, recovery is treated as instantaneous.
    \item \textbf{ReBound-TD-MPC2}: Our method, which replaces the TD-MPC2 critic target with~\eqref{eq:critic-target-rebound}.\footnote{ReBound applies to any off-policy algorithm that learns a Q-function by TD learning (Section~\ref{subsec:sim-sac}); we chose TD-MPC2 over SAC~\cite{haarnoja2018soft} because it learned more stably in preliminary real-vehicle experiments.}
\end{itemize}

\subsubsection{Implementation Details}
We set the discount factor to $\gamma=0.99$ and the observation history length to $N_{\mathrm{h}}=3$.
The LiDAR input consists of 18 equally spaced beams over the full field of view, normalized by 30\,m, so $\myvec{o}_t$ is 28-dimensional and the state is 84-dimensional.
The speed in the reward is normalized by 10\,m/s, and $\lambda_{\mathrm{col}}=10.0$.
A collision is detected when the vehicle position is occupied for three consecutive steps on a local occupancy grid built from the LiDAR point cloud.

The MPPI reset policy maximizes~\eqref{eq:agile-driving-reward} with a horizon of 1.0\,s (20 steps), 3000 samples, $\lambda_{\mathrm{res}}=0.001$, and the KBM as the prediction model; its commanded speed is limited to $\pm1$\,m/s to allow reversing away from walls.
We set $\tau_{\mathrm{m}}=30$.
The MPPI baseline uses a temperature of 0.1 and a maximum speed that varies depending on the environment, with all other settings left unchanged.

For onboard execution, we use a small TD-MPC2 model~\cite{hansen2024tdmpc2} with a latent dimension of 32, two hidden layers of 32 units, four Q-heads, and a planning horizon of 3 (about 40k parameters); the remaining hyperparameters use default values.
The latent dynamics, reward, and termination models and the planner are shared across methods, which differ only in the critic target.
The baseline $\bar{r}$ is initialized to 0 and updated with the mean minibatch TD error using a step size $\eta=0.1$ times the critic learning rate.

\subsection{Simulation Results}
\label{subsec:sim}

\begin{figure*}[t]
    \centering
    \begin{subfigure}[t]{0.32\textwidth}
        \centering
        \includegraphics[width=\linewidth]{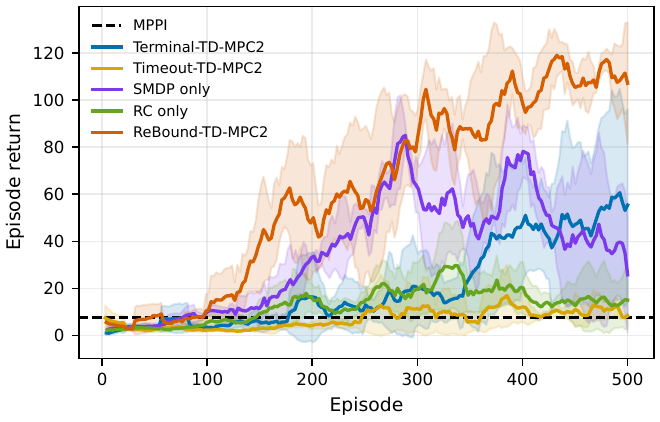}
        \caption{\mytag{LabTrack-Sim}}
        \label{fig:sim_returns_labtrack}
    \end{subfigure}
    \hfill
    \begin{subfigure}[t]{0.32\textwidth}
        \centering
        \includegraphics[width=\linewidth]{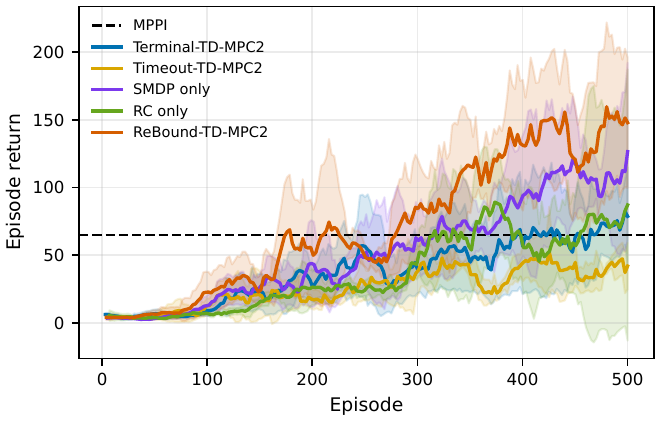}
        \caption{\mytag{MoscowRaceway}}
        \label{fig:sim_returns_moscowraceway}
    \end{subfigure}
    \hfill
    \begin{subfigure}[t]{0.32\textwidth}
        \centering
        \includegraphics[width=\linewidth]{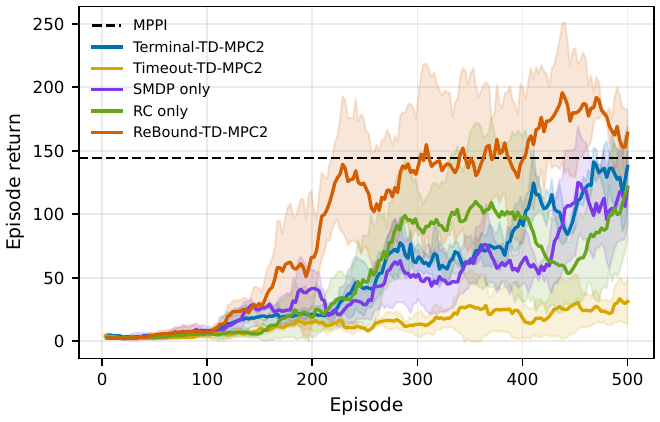}
        \caption{\mytag{Oschersleben}}
        \label{fig:sim_returns_oschersleben}
    \end{subfigure}
    \caption{
        Learning curves on each simulated course.
        Lines show the mean, and bands indicate the standard deviation; dashed lines show the average evaluation return of MPPI.
    }
    \label{fig:sim_returns}
    \vspace{-5mm}
\end{figure*}

\begin{figure}[t]
    \centering
    \begin{subfigure}[t]{\linewidth}
        \centering
        \includegraphics[width=\linewidth]{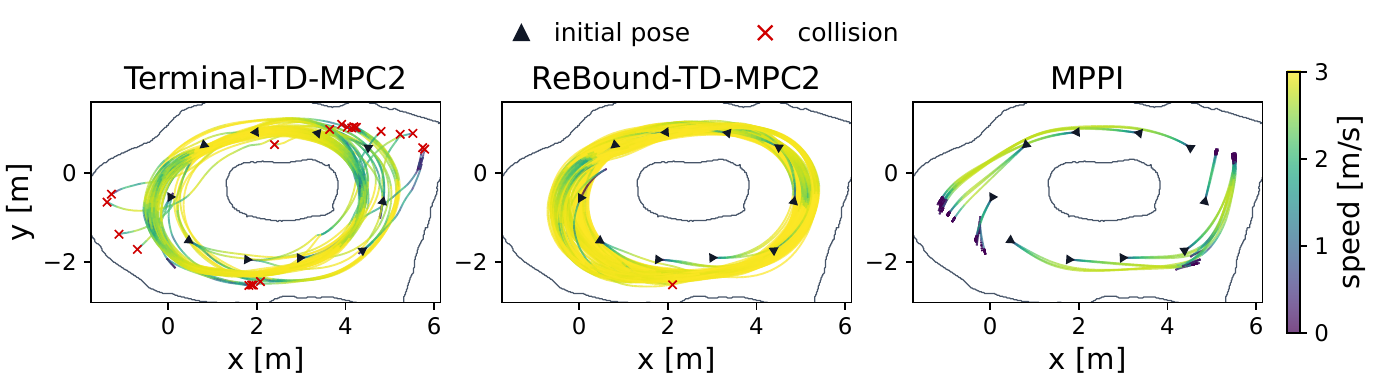}
        \caption{\mytag{LabTrack-Sim}}
        \label{fig:sim_trajectories_labtrack}
    \end{subfigure}

    \begin{subfigure}[t]{\linewidth}
        \centering
        \includegraphics[width=\linewidth]{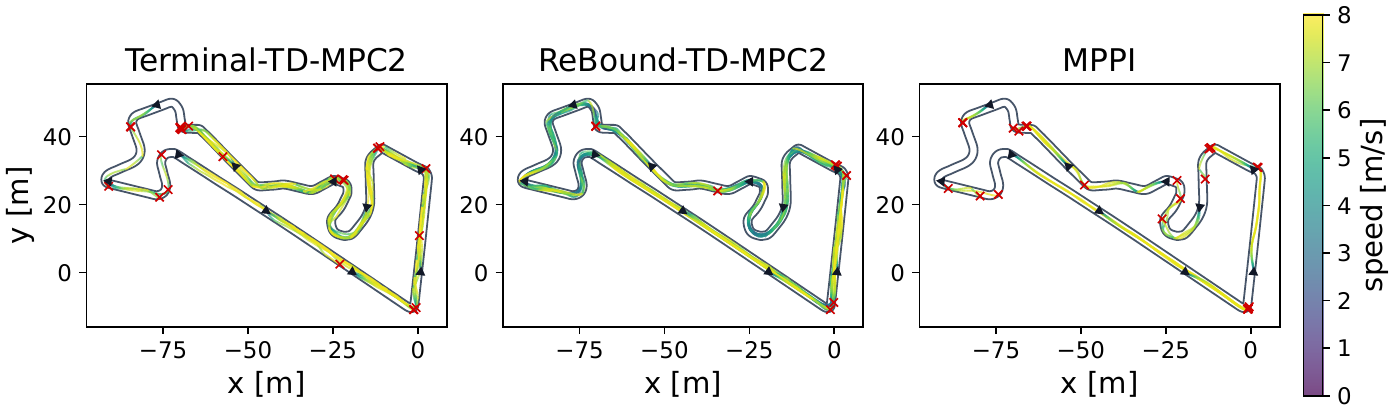}
        \caption{\mytag{MoscowRaceway}}
        \label{fig:sim_trajectories_moscowraceway}
    \end{subfigure}

    \begin{subfigure}[t]{\linewidth}
        \centering
        \includegraphics[width=\linewidth]{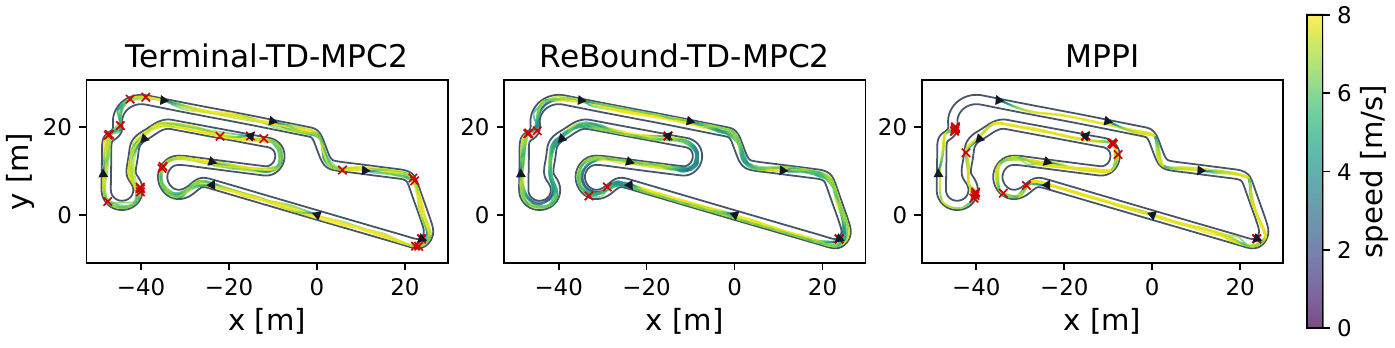}
        \caption{\mytag{Oschersleben}}
        \label{fig:sim_trajectories_oschersleben}
    \end{subfigure}
    \caption{
        Driving trajectories during evaluation on each simulated course (5 seeds $\times$ 10 episodes overlaid).
        Colors indicate speed, black triangles indicate start positions, and red $\times$ symbols mark collisions.
        ReBound-TD-MPC2 approaches the speed limit on straights and decelerates before corners, adapting its driving to each course.
    }
    \label{fig:sim_trajectories}
    \vspace{-3mm}
\end{figure}

\begin{table}[t]
    \centering
    \caption{
        Simulation results (mean $\pm$ standard deviation over 5 seeds $\times$ 10 episodes).
        Italics denote ablations of ReBound-TD-MPC2.
}
    \label{tab:main_results}
    \setlength{\tabcolsep}{3.5pt}
    \renewcommand{\arraystretch}{1.08}

    \begin{tabular}{@{}llcc@{}}
        \toprule
        Environment & Method & Episode Return & Collision Rate \\
        \midrule

        \multirow{6}{*}\mytag{{LabTrack-Sim}}
        & MPPI
        & $7.6 \pm 3.9$
        & $\mathbf{0.00}$ \\
        & Terminal-TD-MPC2
        & $52.9 \pm 64.4$
        & $0.70$ \\
        & Timeout-TD-MPC2
        & $11.1 \pm 4.5$
        & $1.00$ \\
        & ReBound-TD-MPC2
        & $\mathbf{139.7 \pm 10.4}$
        & $0.03$ \\
        \cmidrule(lr){2-4}
        & \quad\textit{SMDP only}
        & $47.4 \pm 59.6$
        & $0.50$ \\
        & \quad\textit{RC only}
        & $18.2 \pm 10.3$
        & $0.23$ \\

        \midrule

        \multirow{6}{*}\mytag{{MoscowRaceway}}
        & MPPI
        & $64.9 \pm 45.5$
        & $1.00$ \\
        & Terminal-TD-MPC2
        & $136.3 \pm 54.2$
        & $0.87$ \\
        & Timeout-TD-MPC2
        & $84.5 \pm 18.7$
        & $1.00$ \\
        & ReBound-TD-MPC2
        & $\mathbf{234.8 \pm 16.1}$
        & $\mathbf{0.30}$ \\
        \cmidrule(lr){2-4}
        & \quad\textit{SMDP only}
        & $130.6 \pm 92.4$
        & $0.97$ \\
        & \quad\textit{RC only}
        & $110.7 \pm 128.8$
        & $0.73$ \\

        \midrule

        \multirow{6}{*}\mytag{{Oschersleben}}
        & MPPI
        & $144.4 \pm 142.4$
        & $0.83$ \\
        & Terminal-TD-MPC2
        & $161.9 \pm 77.5$
        & $0.73$ \\
        & Timeout-TD-MPC2
        & $52.7 \pm 57.3$
        & $1.00$ \\
        & ReBound-TD-MPC2
        & $\mathbf{220.7 \pm 35.5}$
        & $\mathbf{0.33}$ \\
        \cmidrule(lr){2-4}
        & \quad\textit{SMDP only}
        & $83.9 \pm 98.8$
        & $0.93$ \\
        & \quad\textit{RC only}
        & $118.9 \pm 101.3$
        & $0.63$ \\

        \bottomrule
    \end{tabular}
    \vspace{-3mm}
\end{table}

\subsubsection{Driving Performance}
\label{subsec:sim-performance}
Table~\ref{tab:main_results} shows the evaluation results, and Fig.~\ref{fig:sim_returns} shows the learning curves.
ReBound-TD-MPC2 achieved the highest return on all three courses and learned more efficiently than the other learning methods.
It also collided less frequently during training, which matters on real vehicles: the reset policy was active for 10--22\% of the driving time, compared with 22--39\% for Terminal-TD-MPC2.
It also had the lowest evaluation collision rate on all courses except \mytag{LabTrack-Sim}, where MPPI's lower rate results from stopping in front of walls (Fig.~\ref{fig:sim_trajectories_labtrack}).

Timeout-TD-MPC2 performed worst among the learning methods: its return was below that of Terminal-TD-MPC2 on all courses, and all its evaluation episodes ended in collisions.
With time-out bootstrapping, recovery has no cost beyond the explicit collision penalty, so the policy drives too aggressively.

Fig.~\ref{fig:sim_trajectories} shows the evaluation trajectories of the top three methods.
Terminal-TD-MPC2 often collided at corners and did not adequately learn course-specific speed control and racing lines.
MPPI stopped in front of walls on \mytag{LabTrack-Sim} and collided at many corners on the circuits, because its model ignores slip and its finite-horizon optimization lacks a long-term value function, making it prone to local optima.
In contrast, ReBound-TD-MPC2 drove at high speed with almost no collisions on \mytag{LabTrack-Sim}, and on the circuits it decelerated appropriately at corners and chose racing lines suited to each course.

\subsubsection{Ablation Study}
\label{subsec:sim-ablation}
To isolate the contributions of the SMDP formulation and RC, we evaluated \textit{SMDP only}, which uses SMDP bootstrapping without RC, and \textit{RC only}, which adds RC to the collision-terminal update (Fig.~\ref{fig:sim_returns} and Table~\ref{tab:main_results}).
SMDP only was unstable; on \mytag{LabTrack-Sim}, for example, its return rose substantially during training and then declined.
This is consistent with the motivation for RC: long-horizon learning introduces a large common component into the value estimates.
RC only had a lower collision rate than Terminal-TD-MPC2 but converged to slow driving (0.59\,m/s on \mytag{LabTrack-Sim}).
Thus, neither component alone suffices; combining them yields the most stable learning and the best performance.

\begin{figure}[t]
    \centering

    \begin{subfigure}[t]{0.48\linewidth}
        \centering
        \includegraphics[width=\linewidth]{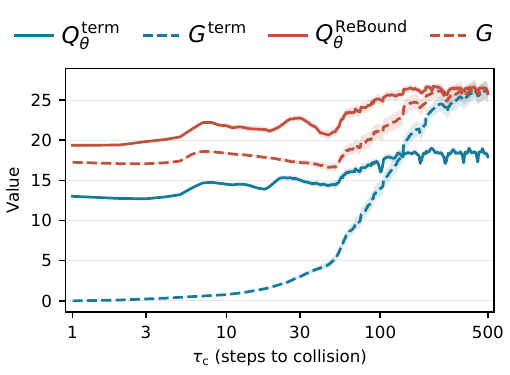}
        \caption{Q-values and learning targets versus the number of steps to collision $\tau_{\mathrm{c}}$.}
        \label{fig:value-analysis-q}
    \end{subfigure}
    \hfill
    \begin{subfigure}[t]{0.48\linewidth}
        \centering
        \includegraphics[width=\linewidth]{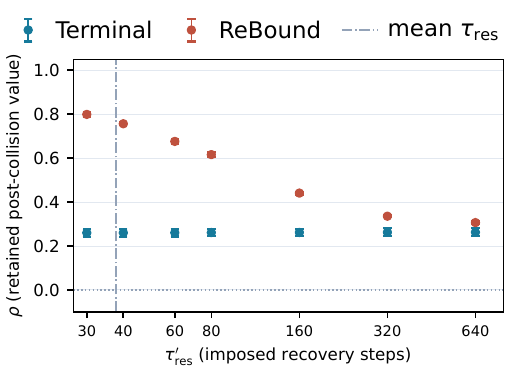}
        \caption{Retained fraction $\rho$ of the post-collision value versus the number of recovery steps $\tau'_{\mathrm{res}}$.}
        \label{fig:value-analysis-rho}
    \end{subfigure}

    \caption{
        Value function analysis on \mytag{LabTrack-Sim}.
        (a) Solid lines: Q-values of each critic; dashed lines: learning targets.
        $Q_\theta^{\mathrm{term}}$ flattens the sharp drop in $G^{\mathrm{term}}$ before a collision, whereas $Q_\theta^{\mathrm{ReBound}}$ follows $G$.
        (b) Mean $\pm$ standard error over 3 seeds; the dash-dotted line marks the observed average recovery length (38 steps).
        ReBound retains a larger fraction of the post-collision value as recovery becomes faster, whereas Terminal does not.
    }
    \label{fig:value-analysis}
    \vspace{-3mm}
\end{figure}

\subsubsection{Value Function Analysis}
\label{subsec:sim-value}
\paragraph{Number of steps to collision $\tau_{\mathrm{c}}$ and Q-value}
We compare the critics trained on \mytag{LabTrack-Sim} by Terminal-TD-MPC2 and ReBound-TD-MPC2 (hereafter Terminal and ReBound) using identical data: 3,456 rollouts of 1,000 steps from the Terminal policy, with 8 per combination of 108 random start states and 4 initial actions.
For each state-action pair, we computed the discounted returns of the two learning targets ($G^{\mathrm{term}}$ in~\eqref{eq:q-term} and $G$ in~\eqref{eq:return-rf}), as well as the Q-values ($Q_\theta^{\mathrm{term}}$, $Q_\theta^{\mathrm{ReBound}}$) of the respective trained critics\footnote{The Q-values of ReBound were converted back to the pre-RC scale using~\eqref{eq:rc-shift}.}.

Fig.~\ref{fig:value-analysis-q} plots these quantities against $\tau_{\mathrm{c}}$.
Even on the same data, the two learning targets have very different shapes.
$G^{\mathrm{term}}$ drops to almost zero just before a collision, whereas the drop in $G$ is only about one third of its value far from collisions (about 26).
This difference corresponds to the mismatch in~\eqref{eq:terminal-bias}, \ie, the hidden collision penalty.

Next, we examine how well each critic represents its own target.
$Q_\theta^{\mathrm{term}}$ is flatter than $G^{\mathrm{term}}$: the Terminal critic fails to represent the sharp drop just before a collision, precisely where accuracy matters most in agile driving.
In contrast, because ReBound bootstraps from the state after recovery, its target $G$ exhibits only a shallow dip, and its critic follows $G$ more faithfully.
That is, ReBound is not a more efficient Q-learner per se; its target is simply easier to learn, so the critic represents the value more accurately.

\paragraph{Number of recovery steps $\tau'_{\mathrm{res}}$ and Q-value}
\label{subsec:sim-recovery}

Section~\ref{subsec:terminal-comparison} predicts that the gap between ReBound and collision-terminal values grows as recovery becomes faster.
To verify this, we changed only the recovery length for each collision in these rollouts to $\tau'_{\mathrm{res}}\in\{30,40,60,80,160,320,640\}$,
and retrained small critics (two layers of 32 units, without RC, 3 seeds) with respective targets $G^{\mathrm{term}}$ and $G$, using identical minibatches and initial weights.
For 374 collisions from start states held out from critic training, we compared the fraction of the post-collision value retained in the Q-value of the collision-causing action,
\begin{align}
    \rho
    =
    \frac{
        \mathrm{mean}\left(Q_\theta(\myvec{s}_{t_{\mathrm{c}}},\myvec{a}_{t_{\mathrm{c}}})-r_{t_{\mathrm{c}}}\right)
    }{
        \mathrm{mean}\left(\gamma\,G_{t_{\mathrm{rec}}}\right)
    },
\end{align}
where $G_{t_{\mathrm{rec}}}$ is the return~\eqref{eq:return-rf} after recovery, common to both methods, and $Q_\theta\in\{Q_\theta^{\mathrm{term}},Q_\theta^{\mathrm{ReBound}}\}$ is the critic trained with the respective target.
Ideally, $\rho=0$ for Terminal and $\rho\approx\gamma^{\tau'_{\mathrm{res}}}$ for ReBound.

As shown in Fig.~\ref{fig:value-analysis}(b), the value of $\rho$ for Terminal was almost constant regardless of $\tau'_{\mathrm{res}}$, whereas that of ReBound decreased from 0.80 at $\tau'_{\mathrm{res}}=30$ to 0.31 at $\tau'_{\mathrm{res}}=640$, retaining the post-collision value discounted by the recovery time.
This confirms that the gap between the two grows as recovery becomes faster: our setting, where the vehicle recovers within a few seconds, is precisely where the collision-terminal target deviates most from the reset-free objective.

\subsubsection{Application to Another Off-Policy Algorithm}
\label{subsec:sim-sac}
Applied to SAC~\cite{haarnoja2018soft}, ReBound improved the average evaluation return on two of the three courses
(\mytag{LabTrack-Sim}: 7.1 $\rightarrow$ 14.1, \mytag{MoscowRaceway}: 177.4 $\rightarrow$ 185.6).
On \mytag{Oschersleben}, however, the return decreased (256.5 $\rightarrow$ 186.8).
Since ReBound changes only the critic target, its effect depends on how the base algorithm uses the critic: SAC's actor directly follows the critic's gradient, whereas TD-MPC2 uses an ensemble critic only as the terminal value of short-horizon planning.
A detailed analysis of this interaction is left for future work.

\subsection{Real-World Results}
\label{subsec:real}

\begin{figure*}[t]
    \centering
    \includegraphics[width=\textwidth]{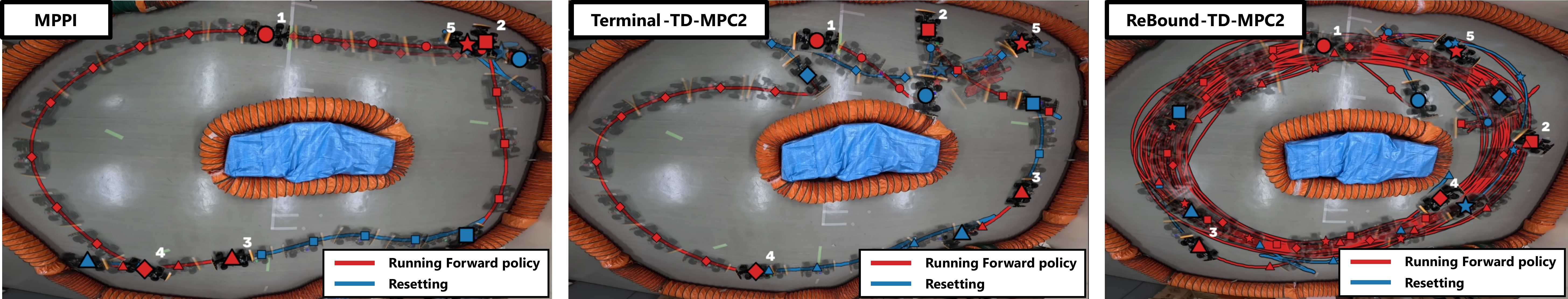}
    \caption{
        Driving after training on the real \mytag{LabTrack} with the reset policy enabled.
        Markers distinguish episodes, and numbers indicate start positions.
}
    \label{fig:real-eval-traj}
    \vspace{-4mm}
\end{figure*}

\begin{figure}[t]
\centering
\includegraphics[width=0.9\linewidth]{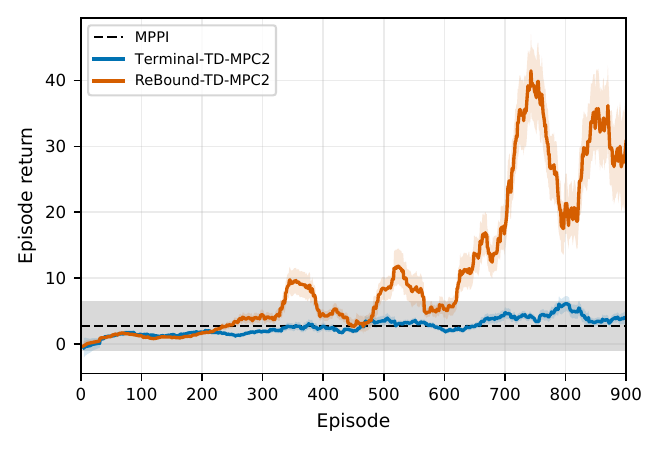}
\caption{
    Episode returns during training on the real \mytag{LabTrack}.
    Moving average over 50 episodes; bands show the standard error within the window.
    The dashed line and gray band show the mean and standard deviation of MPPI's evaluation returns.
}
\label{fig:reward-curve-lab-track}
\vspace{-4mm}
\end{figure}

\begin{table}[t]
\centering
\caption{
    Evaluation results over 10 episodes on the real \mytag{LabTrack} (mean $\pm$ standard deviation).
    Without manual resets: the reset policy recovers the vehicle, and driving continues for 500 steps.
    With manual resets: a person positions the vehicle, and the episode ends at the first collision.
}
\label{tab:real-world-evaluation}
\small
\setlength{\tabcolsep}{5pt}
\begin{tabular}{lrr}
\toprule
Method & Episode Return & Average Speed (m/s) \\
\midrule
\multicolumn{3}{l}{\textit{Without manual resets (continuing)}} \\
MPPI
    & $1.67 \pm 2.74$
    & $0.40 \pm 0.60$ \\
Terminal-TD-MPC2
    & $3.89 \pm 3.62$
    & $1.36 \pm 0.55$ \\
ReBound-TD-MPC2
    & {$\mathbf{68.26 \pm 56.12}$}
    & {$\mathbf{2.00 \pm 0.66}$} \\
\midrule
\multicolumn{3}{l}{\textit{With manual resets (ends at first collision)}} \\
MPPI
    & $8.62 \pm 3.83$
    & $1.56 \pm 0.38$ \\
Terminal-TD-MPC2
    & $2.19 \pm 1.74$
    & $1.22 \pm 0.47$ \\
ReBound-TD-MPC2
    & {$\mathbf{73.33 \pm 60.96}$}
    & {$\mathbf{2.01 \pm 0.54}$} \\
\bottomrule
\end{tabular}
\end{table}

\subsubsection{Comparison on \mytag{LabTrack}}
On the real \mytag{LabTrack}, we trained Terminal-TD-MPC2 and ReBound-TD-MPC2 once each and evaluated three methods, including MPPI, from 10 start positions (Fig.~\ref{fig:overview}, right: ReBound-TD-MPC2 at the start of training and after about 400 and 800 episodes).

The return of ReBound-TD-MPC2 improved substantially after about 600 episodes (Fig.~\ref{fig:reward-curve-lab-track}), and in the latter half of training, it completed laps continuously (Fig.~\ref{fig:real-eval-traj}).
In contrast, the return of Terminal-TD-MPC2 improved only slightly, and it continued to collide and recover even late in training.
MPPI also collided frequently because it cannot account for slip.
As Table~\ref{tab:real-world-evaluation} shows, only ReBound-TD-MPC2 achieved a markedly high return and average speed, both when driving continued through recoveries as in training (Fig.~\ref{fig:real-eval-traj}) and when episodes ended at the first collision.
These results demonstrate that ReBound efficiently learns high-performance driving in the real world without manual resets.

\subsubsection{Continual Learning on \mytag{LabCircuit}}
We continued training on \mytag{LabCircuit}, starting from the ReBound-TD-MPC2 policy trained on \mytag{LabTrack} (see the supplementary video).
During evaluation from 10 start positions, ReBound-TD-MPC2 drove $20.5\pm20.8$\,m per episode at $1.93\pm0.66$\,m/s, whereas MPPI drove $5.6\pm7.3$\,m at $0.86\pm0.92$\,m/s.
Thus, even on a new course, ReBound substantially outperformed MPPI.

%% file: src/conclusion.tex
\section{Conclusion}
\label{sec:conclusion}

We presented ReBound for reset-free RL in real-vehicle agile driving.
It applies SMDP bootstrapping by treating a collision-causing action and subsequent recovery as a single transition.
Simulation and real-world experiments showed that the conventional collision-terminal update is misaligned with the reset-free objective and introduces a hidden collision penalty absent from the reward function.
Only ReBound achieved stable learning and high driving performance.

This work has several limitations.
First, the reset policy is a fixed, hand-designed MPPI controller, and collisions are assumed to be recoverable.
Jointly learning the reset and forward policies would make the reset interval non-stationary from the forward policy's perspective.
Extending ReBound to this setting and to irrecoverable failures remains future work.
Second, the interaction between ReBound and the base learner, as observed with SAC, requires further analysis.